\documentclass[11pt]{article}
\usepackage[normalem]{ulem}

\usepackage[final]{acl}

\usepackage{times}
\usepackage{latexsym}
\usepackage{booktabs}
\usepackage{tabularx}
\usepackage{array}
\usepackage[T1]{fontenc}
\usepackage[utf8]{inputenc}
\usepackage{easyReview}
\usepackage{enumitem}
\usepackage{multirow}
\usepackage{graphicx}
\usepackage{algorithm}
\usepackage{algpseudocode}
\usepackage{amsmath} % For mathematical symbols
\usepackage{microtype}

\usepackage{inconsolata}
\usepackage{caption}

\usepackage[most]{tcolorbox}
\usepackage{xcolor}
\usepackage{enumitem}
\usepackage{ragged2e}
\usepackage{microtype}

\definecolor{CaseBlue}{HTML}{315A7D}
\definecolor{CaseBlueLight}{HTML}{EEF5FA}
\definecolor{CaseGold}{HTML}{B7791F}
\definecolor{CaseGoldLight}{HTML}{FFF8E8}
\definecolor{CaseGreen}{HTML}{34745A}
\definecolor{CaseGreenLight}{HTML}{EFF8F3}
\definecolor{CaseGray}{HTML}{5B6573}
\definecolor{CaseGrayLight}{HTML}{F5F6F8}

\tcbset{
  turnbox/.style={
    enhanced,
    breakable,
    colback=CaseBlueLight,
    colframe=CaseBlue,
    colbacktitle=CaseBlue,
    coltitle=white,
    fonttitle=\bfseries,
    boxrule=0pt,
    leftrule=2.2pt,
    arc=1.5mm,
    left=3mm,
    right=3mm,
    top=2mm,
    bottom=2mm,
    before skip=7pt,
    after skip=7pt
  },
  querybox/.style={
    enhanced,
    breakable,
    colback=CaseGrayLight,
    colframe=CaseGray,
    boxrule=0pt,
    leftrule=2.2pt,
    arc=1.5mm,
    left=3mm,
    right=3mm,
    top=2mm,
    bottom=2mm,
    before skip=7pt,
    after skip=7pt
  },
  resultbox/.style={
    enhanced,
    breakable,
    colback=CaseGreenLight,
    colframe=CaseGreen,
    boxrule=0pt,
    leftrule=2.2pt,
    arc=1.5mm,
    left=3mm,
    right=3mm,
    top=2mm,
    bottom=2mm,
    before skip=7pt,
    after skip=7pt
  },
  strategybox/.style={
    enhanced,
    breakable,
    colback=CaseGoldLight,
    colframe=CaseGold,
    colbacktitle=CaseGold,
    coltitle=white,
    fonttitle=\bfseries,
    boxrule=0pt,
    leftrule=2.2pt,
    arc=1.5mm,
    left=3mm,
    right=3mm,
    top=2mm,
    bottom=2mm,
    before skip=7pt,
    after skip=7pt
  }
}

\usepackage{graphicx}

\title{Clarify User Expertise: Towards Proactive Conversational Agents \\Tailoring Responses to User Proficiency}

\newcommand{\ours}{\textsc{Passing}}

\author{
Zhihong Cao$^{1}$\quad  Chen Huang$^{2,3}$\thanks{Corresponding author.} \\
$^{1}$ School of Computing and Data Science, The University of Hong Kong  \\
$^{2}$ Institute of Data Science, National University of Singapore  \\
$^{3}$ College of Computer Science, Sichuan University  \\
\texttt{czh040302@163.com, huangc.scu@gmail.com}
}

\begin{document}
\maketitle
\begin{abstract}
In the context of information seeking, conversational agents are undergoing an evolution from reactive tools to proactive, personalized assistants. A critical aspect of this evolution is the ability to tailor strategic interactions to a user's unique needs and expectations. 
Unlike existing studies that focus on proactively clarifying query ambiguities, we center on clarifying the user's expertise in order to tailor responses for better user comprehension.
We find that existing agents struggle to determine user expertise from queries alone, a limitation that prevents them from dynamically adapting their responses.
To address this gap, we introduce \ours~to empower the agent to proactively clarify a user's expertise through targeted inquiries. This is achieved by our \textit{What-to-Ask} and \textit{How-to-Ask} strategies, induced by LLM self-play. Our extensive experiments also show our superiority. We believe that \ours~represents a crucial step towards creating more human-centric conversational agents.
\end{abstract}

\section{Introduction}
Conversational agents have become a crucial tool for satisfying user information needs in multi-turn dialogues \cite{casheekar2024contemporary, zamani2023conversational}. With the advent of large language models (LLMs), a new paradigm of \textit{Proactive Conversational Agents} has emerged that goes beyond simply reacting to user queries \cite{deng2025proactive, liao2023proactive, huang-etal-2026-towards}. These agents take the initiative to handle under-specified \cite{braslavski2017you, xu-etal-2019-asking} or over-specified user requests \cite{wu-etal-2023-inscit, min-etal-2019-multi} during the information seeking process, rather than passively following the user's lead. Common proactive behaviors include asking questions to clarify query ambiguity \cite{zhang-etal-2024-clamber, chen-etal-2024-style} or elicit user preference \cite{10.1145/3269206.3271776, 10.1145/3485447.3512088, chen2023travel}. For instance, in response to a broad query like "quantum computing", these agents might ask clarifying question: "Are you asking about the history of quantum computing?". As such, the agent seeks to proactively comprehend user query to deliver a more accurate response.

\begin{figure}
    \centering
    \includegraphics[width=0.38\textwidth]{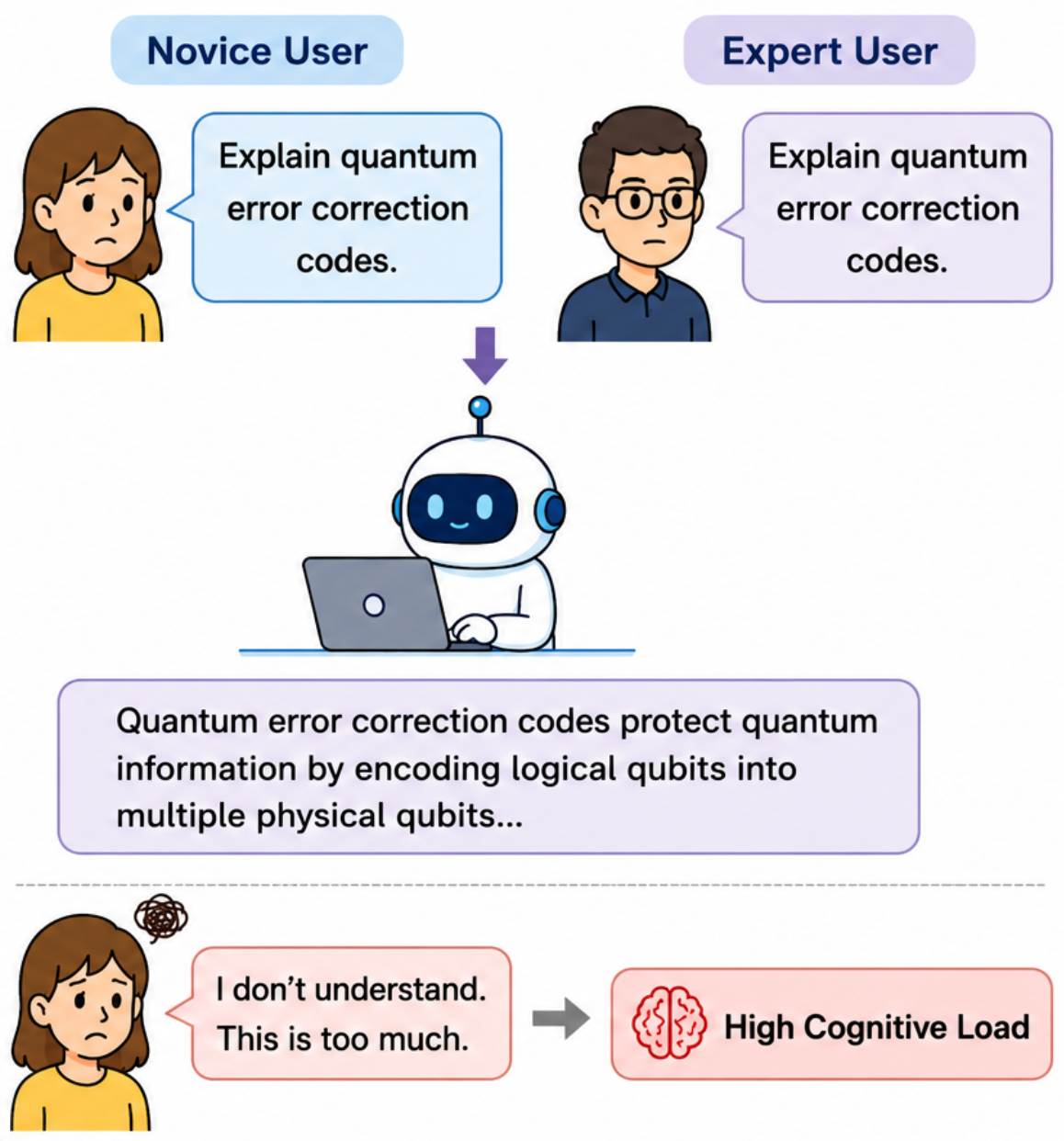}
    \caption{User expertise can vary depending on the specific query. LLM agent should tailor its response based on user expertise.}
    \label{fig:demo}
    % \vspace{-3mm}
\end{figure}

While current research prioritizes response accuracy, the crucial task of tailoring responses for user comprehension has been largely overlooked. Modern users expect responses that are not only factually correct but also personalized to their individual domain expertise, adjusting the level of detail and complexity accordingly \cite{dreyfus1980five, schank1999dynamic, chen2024learning, yao-etal-2024-readme, 10.1145/3706599.3719982, tawfik2021exploring, yaghoubzadeh-kopp-2017-enabling}. As illustrated in Figure \ref{fig:demo}, user expertise can vary significantly depending on the specific query. LLM agents must estimate the user's query-specific expertise level prior to formulating a response. Otherwise, novices, for example, can be overwhelmed by technical details, making them susceptible to cognitive overload \cite{dreyfus1980five}.

To this end, our work centers on clarifying user expertise as a foundational step for generating tailored responses\footnote{Our research scope is confined to the task of query-specific user expertise estimation, a prerequisite for generating tailored responses. Response features required by experts and novices is provided in Appendix \ref{expert}.}. We begin by experimentally investigating the limitations of current agents in query-specific expertise estimation (Section \ref{preE}). Our findings reveal a significant expertise inconsistency in LLM-based agents: when presented with the same query, their predictions of user expertise are highly stochastic, with high inconsistency rate. This aligns with previous user study \cite{palta2025speaking} and implies that a user's expertise on the specified query is difficult to infer from the query text \cite{10.1145/3616855.3635845}. For instance, a computer science PhD student may be a novice in physics but inquiry technical terms like 'quantum computing' without a full grasp of the concept. Therefore, we argue that \textbf{proactive conversational agents must probe the user's expertise on a given query via strategic interaction}, a need previously identified \cite{chen2024learning} but not yet fulfilled.

In this paper, we propose \textbf{\ours}, the first method for \uline{P}ro\uline{A}ctive u\uline{S}er experti\uline{S}e prob\uline{ING} through targeted inquiries. Recognizing that expertise is often manifested in a user's knowledge scope and reasoning processes \cite{feltovich1997expertise, palta2025speaking}, \ours~employs a strategy-guided probing mechanism, which utilizes LLM-induced strategies (i.e., \textit{What-to-Ask} and \textit{How-to-Ask}) to direct \ours~in formulating informative inquiries. As illustrated in Figure \ref{fig:main2}, \ours~leverages these strategies to iteratively pose targeted questions at each turn, obtaining user information on the knowledge scope and reasoning logic based on user's responses. To balance the trade-off between maximizing information gain and minimizing conversational overhead, \ours~stops probing after a pre-defined number of turns. Subsequently, it estimates the user's query-specific expertise and generates an appropriately tailored response.
Before putting \ours~into use, \ours~implements an offline self-play simulation to derive these strategies: dialogue histories between \ours~and simulated users are analyzed to extract effective probing strategies, which are then iteratively optimized throughout the self-play process. Notably, distinct from research on ambiguity resolution or preference elicitation, \ours~prioritizes the assessment of user expertise to facilitate better answer comprehension, addressing a vital gap in human-centric AI \cite{deng2024towards}.

To evaluate the effectiveness of \ours, we conduct experiments with various baselines and LLM backbones using various datasets. Experimental results demonstrate that with an average of only 1.2 inquiry turns, \ours~achieves nearly a threefold improvement in the accuracy of user expertise estimation. Finally, we sum up our main contributions as follows.
\begin{itemize}[leftmargin=*]%,itemindent=0.05cm, itemsep=-2pt]
    \item We highlight the importance of user expertise estimation for proactive conversational agents, a step toward more human-centric interactions.
    \item We propose \ours, featuring accurate expertise estimation through targeted inquiries while minimizing conversational overhead.
    \item We experimentally demonstrate that existing agents fail to reliably infer user expertise from queries, and validate our effectiveness.% of \ours, with in-depth analysis revealing \add{the critical role of the XX mechanism}.
\end{itemize}

\section{Related Work}

\noindent\textbf{User Expertise Estimation}. A body of theoretical \cite{dreyfus1980five, schank1999dynamic} and user-centric \cite{chen2024learning, schubert2013characterizing, palta2025speaking} research highlights the significant differences between novices and experts in their perceptions and needs. 
In this case, to generate such personalized response, existing research largely assumes that user expertise is a known information (e.g., via pre-defined user profiles) \cite{10.1145/3706599.3719982, sun2025persona, cheng-etal-2024-dialogues}. However, such information is insufficient for determining query-specific expertise, because a user's expertise can vary significantly from one query to the next. While recent methods like ExpertPrompting \cite{xu2025expertpromptinginstructinglargelanguage} attempt to infer expertise via in-context learning, user studies indicate that LLM-based agents frequently miscalibrate—either underestimating or overestimating the user's knowledge—resulting in reduced satisfaction \cite{palta2025speaking}. Therefore, proactive agents must be able to probe for this information before generating a response. To fill this gap, we propose \ours, the first method designed to leverage an agent’s proactivity to probe and clarify a user's expertise on the user query through targeted inquiries.

\noindent\textbf{Proactive Conversational Agents}. Analogous to general proactive agents that explore an environment to acquire information that improves future decision-making \cite{lu2025proactive, guan2026clearingfoginstallingrefining}, proactive conversational agents actively engage users to uncover task-relevant information and facilitate goal completion. These agents are distinguished by their ability to take initiative, steering dialogues toward productive outcomes rather than passively following a user's lead \cite{deng2024towards, deng2025proactive}. To date, proactive efforts in information seeking have primarily focused on improving response accuracy. This is achieved through strategies such as clarifying query ambiguity \cite{zhang-etal-2024-clamber, chen-etal-2024-style}, eliciting user preferences \cite{10.1145/3269206.3271776, 10.1145/3485447.3512088, chen2023travel}, or managing over-specified requests \cite{wu-etal-2023-inscit, min-etal-2019-multi}. However, this intense focus on accuracy has left the crucial task of tailoring responses for user comprehension largely overlooked. User studies indicate a clear demand for agents that first assess a user's knowledge level before providing an answer \cite{chen2024learning}. In response to this need, our work introduces a proactive conversational agent designed specifically to probe for a user's expertise on a given query before generating a response.

\subsection{Prediction Stability of Existing Methods}

\begin{table}[h]
\centering
\resizebox{0.45\textwidth}{!}{%
\begin{tabular}{l|cc|cc}
\toprule
\multirow{2}{*}{\textbf{Method}} & \multicolumn{2}{c|}{\textbf{MMLU-Pro}} & \multicolumn{2}{c}{\textbf{ARC-MCAS}} \\ \cline{2-5}
 & \textbf{Kappa} & \textbf{Unk} & \textbf{Kappa} & \textbf{Unk} \\ \midrule
Deterministic      & 100.0 & 0.00  & 100.0 & 0.00  \\
Fully Random       & 0.00  & 16.67 & 0.00  & 16.67 \\ \midrule
Direct Prompt      & 41.25 & 28.00 & 30.64 & 82.80 \\
CoT                & 28.47 & 26.40 & 70.34 & 89.60 \\
Self-consistency   & 85.12 & 32.00 & 75.35 & 86.80 \\
Diverse-Aspect     & 39.91 & 14.80 & 47.23 & 25.60 \\ \midrule
IDL*               & 40.31 & 23.79 & 15.00 & 4.40  \\
ExpertPrompting    & 36.25 & 25.60 & 72.00 & 94.00 \\
\bottomrule
\end{tabular}%
}
\caption{Prediction stability and unknown rate of existing methods for query-specific user expertise estimation.}
% \vspace{-3mm}
\label{tab:preE}
\end{table}

\begin{figure*}
    \centering
    \includegraphics[width=0.98\textwidth]{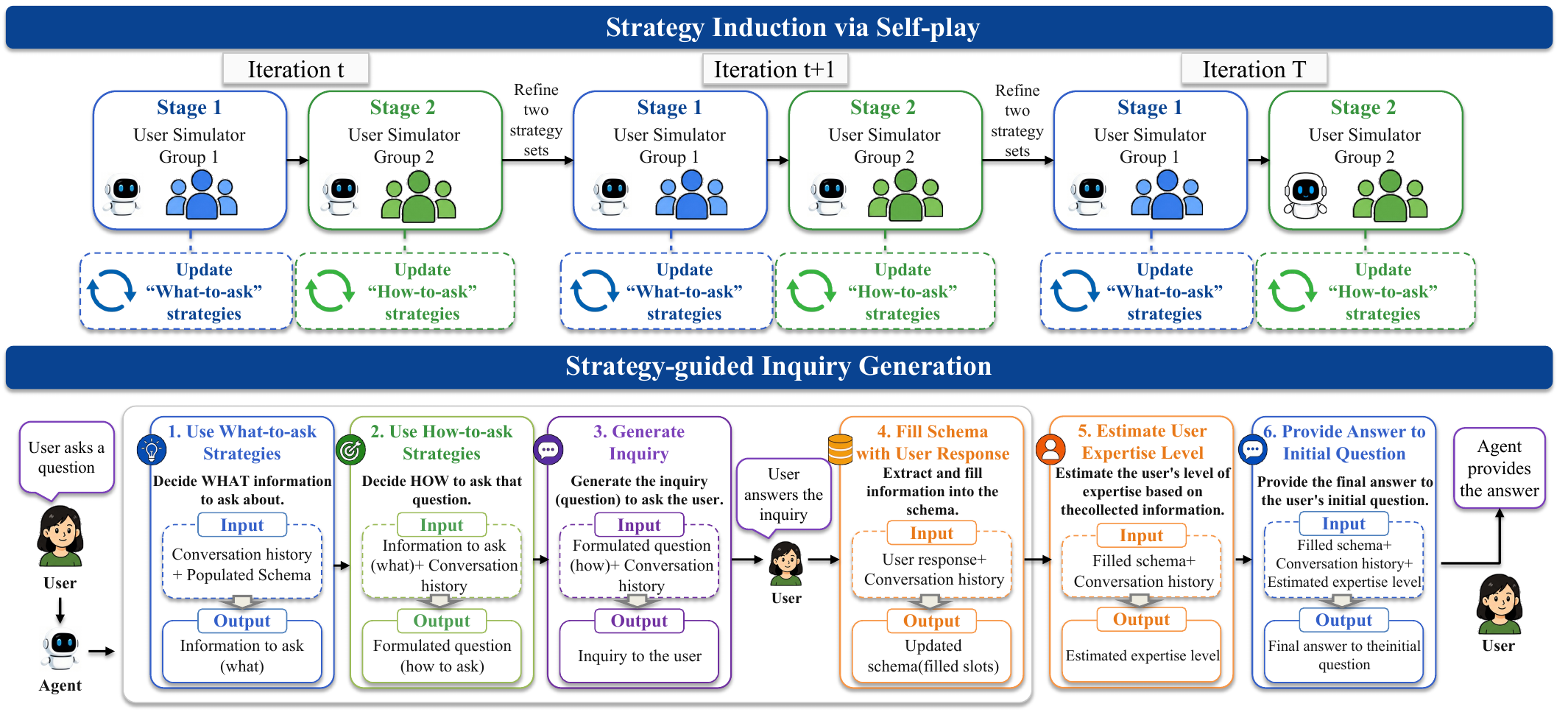}
    \caption{Overview of \ours. It utilizes LLM-induced strategies (i.e., What-to-Ask and How-to-Ask), obtained through self-play simulation.}
    \label{fig:main2}
    % \vspace{-3mm}
\end{figure*}

\section{Preliminary Experiment}
\label{preE}
This section investigates LLM-based agents can reliably determine user expertise on a given query.

\subsection{Experiment Setup}
\noindent\textbf{Experiment Overview}. A single query can be posed by users with vastly different expertise levels. For instance, a query about \textit{quantum mechanics} could originate from a physics novice, a physics expert, or even a computer science expert who is a novice in quantum mechanics. Considering the scarcity of open-source data labeled with query-specific user expertise, we propose an evaluation methodology focused on \uline{prediction stability} of an agent's expertise judgments when it is presented with the same query multiple times.

\noindent\textbf{Baselines}. As there are no established methods specifically for probing query-specific user expertise, our baselines include two distinct categories. 1) \textbf{General LLM-based Agents}: \uline{Direct Prompt}, \uline{Chain-of-Thought} (CoT) \cite{wei2023chainofthoughtpromptingelicitsreasoning}, \uline{Self-consistency} \cite{wang2022self}, and our designed \uline{Diverse-Aspect} that first prompts the LLM to generate judgments from multiple specific aspects (e.g., terminology use) and then ensembles these aspect-specific judgments to produce the final prediction. 2) \textbf{Expertise Estimation}: \uline{IDL} \cite{cheng-etal-2024-dialogues} and \uline{ExpertPrompting} \cite{xu2025expertpromptinginstructinglargelanguage}. Notably, IDL predicts general user expertise based solely on pre-collected user conversation history\footnote{History is built from validation set of corresponding data.} and are inherently unable to provide an estimation specific to the given input query. To the best of our knowledge, ExpertPrompting is the only query-specific expertise estimation method. Finally, GPT-4o serves as the LLM backbone.

\noindent\textbf{Dataset}. To ground our evaluation in realistic and challenging scenarios, we use questions from two multi-disciplinary datasets as input user queries: MMLU-Pro \cite{wang2024mmlu} and ARC-MCAS\footnote{Subset of the ARC-Challenge} \cite{clark2018think}. For the output classes, we adopt the \textit{Dreyfus Model} \cite{dreyfus1980five}, which categorizes user expertise into five levels: \uline{Novice}, \uline{Advanced Beginner}, \uline{Competent}, \uline{Proficient}, and \uline{Expert} (detailed in Appendix \ref{expert}). We further augment these five expertise levels with an additional `\textit{Unknown}' category. This results in a six-way classification task where each baseline must classify the user's expertise on the input query into one of these categories. A more comprehensive evaluation is presented in Section \ref{exp}.

\noindent\textbf{Evaluation Metrics}. Given the multi-class classification nature of our task, we measure prediction stability using \uline{Fleiss' Kappa} \cite{fleiss1971measuring}. This metric measures the inter-run agreement for individual predictions. We treat the set of predictions generated under each random seed as an independent "rater" and calculate the Kappa score across all user queries to assess consistency between runs. Additionally, we calculate the proportion of `\uline{Unknown}' predictions generated by the model, which serves as a direct statistical measure of the model's self-assessed inability to determine user query-specific expertise.
Refer to Appendix \ref{metrics} for details.

\noindent\textbf{Implementation Details}. 
To evaluate the stability of each baseline's judgment, we conducted 5 independent runs for every query. Each of these 5 runs was initialized with the same pre-defined random seeds. More details are in Appendix \ref{impl}.

\subsection{Results}

\noindent\textbf{LLM agents are unreliable for robustly estimating user expertise specific to a given query}. 
As illustrated in Table~\ref{tab:preE}, while existing methods demonstrate improved stability compared to \textit{fully random} approaches when assessing user query-specific expertise, they still lack consistency compared to the \textit{deterministic} method, often fluctuating in their judgments. The consistently low Kappa scores across methods reveal that even when models do produce a prediction, their judgments are highly inconsistent across runs. Notably, while \textit{Self-consistency} achieves a relatively high Kappa (85.12 on MMLU-Pro, 75.35 on ARC-MCAS), this stability is an artifact of its majority-voting mechanism, which is the aggregation of multiple samples naturally converges to a dominant prediction, rather than reflecting genuine expertise inference. Similarly, the high Kappa of certain methods on ARC-MCAS (e.g., \textit{CoT}: 70.34, \textit{ExpertPrompting}: 72.00) is largely driven by an extremely high Unknown rate (89.60\% and 94.00\% respectively), indicating that the model repeatedly abstains rather than making meaningful predictions. Importantly, \textit{ExpertPrompting}, which is designed to estimate general user expertise, typically fails to accurately predict query-level expertise. This highlights that any user can copy-paste a jargon-heavy query from the web into an LLM, creating a façade of expertise despite having no actual proficiency in the domain. This motivates our proactive agent to probe the user's query-level expertise.

\section{\ours: Clarify User Expertise}

\noindent\textbf{Overview}. As illustrated in Figure~\ref{fig:main2}, \ours~estimates a user's query-specific expertise through iterative probing and slot-based state tracking. Given a user query $q$, \ours~first constructs a query context $C_0$ using the query and model-generated answer. It then engages the user in a multi-turn probing process guided by predefined probing strategies $S$. Formally, let $H_t={(p_i,u_i)}_{i=1}^{t-1}$ denote the conversation history before turn $t$, where $p_i$ and $u_i$ represent the system inquiry and user response, respectively, with $u_1=q$. Let $L_t$ denote the current slot-based expertise schema. At each turn, \ours~generates an inquiry $p_t$ based on the current state $(C_0,L_t)$, receives the user's response $u_t$, and updates both the conversation state and schema to incorporate newly acquired evidence about the user's expertise. This process repeats until a termination condition is met. Then \ours~produces a query-specific expertise assessment and generates a personalized response accordingly.

\noindent\textbf{Multi-turn Probing and Slot Filling}.
To capture query-specific expertise, we design an expertise schema based on the Dreyfus Model \cite{dreyfus1980five, mangiante2021teaching}, where each slot corresponds to a key assessment aspect (Table~\ref{tab:schema}). Following prior work on dialogue state tracking \cite{cao-etal-2025-breaking, das2024s3}, \ours~adopts a slot-filling paradigm that incrementally populates the schema using evidence collected during probing. After each user response, relevant slots are updated to reflect newly inferred expertise signals, and the updated schema is subsequently used to guide the next inquiry. To further improve inquiry generation, we incorporate LLM-induced probing strategies, described in Section~\ref{inquiry_gen}.

\noindent\textbf{Query-specific Expertise Assessment}.
The probing process terminates when either the maximum probing budget $T$ is reached or all schema slots have been populated. \ours~then estimates the user's query-specific expertise using the populated schema together with the accumulated conversation history, and generates a response tailored to both the user's query and estimated expertise level.

\begin{table*}[]
\centering
\resizebox{0.97\textwidth}{!}{%
\begin{tabular}{l|l}
\toprule
\textbf{Aspects} & \textbf{Descriptions} \\ \midrule
\begin{tabular}[c]{@{}l@{}}Knowledge Scope\\ (What they know)\end{tabular} & \begin{tabular}[c]{@{}l@{}}The extent, type, and organization of the knowledge of the person: \\ knowing ``facts'' (explicit knowledge), knowing when to apply it (applied knowledge), \\ or knowing ``how'' intuitively (tacit knowledge)\end{tabular} \\ \midrule
\begin{tabular}[c]{@{}l@{}}Components \\ (What they perceive)\end{tabular} & \begin{tabular}[c]{@{}l@{}}What types of knowledge or information the person relies on: \\ context-free (rules, abstract principles), rule-based (context-aware rule selection), \\ or situational (contextual, experiential patterns)\end{tabular} \\ \midrule
\begin{tabular}[c]{@{}l@{}}Perspective\\ (How they organize info)\end{tabular} & \begin{tabular}[c]{@{}l@{}}How the person filters and prioritizes information: overwhelmed easily, \\ making a deliberate plan to filter noise, or instinctively seeing important information without trying\end{tabular} \\ \midrule
\begin{tabular}[c]{@{}l@{}}Action\\ (How they decide)\end{tabular} & How decisions are made: analytic (step-by-step reasoning) or intuitive (immediate, holistic response) \\ \midrule
\begin{tabular}[c]{@{}l@{}}Commitment\\ (Emotional connection)\end{tabular} & \begin{tabular}[c]{@{}l@{}}The level of personal agency and emotional investment in the outcome: \\ detached (disengaged, rule-following) or involved (personally invested, responsible for outcome)\end{tabular} \\ \bottomrule
\end{tabular}%
}
\caption{Structured schema used for estimating user expertise based on the Dreyfus Model.}
% \vspace{-3mm}
\label{tab:schema}
\end{table*}

\subsection{Strategy Induction via Self-play}
\label{induction}
The key challenge in expertise estimation lies in generating probing questions that effectively reveal user expertise while minimizing cognitive burden. This requires determining both \textit{what} to ask and \textit{how} to ask. To this end, \ours~learns these strategies through offline self-play with user simulators\footnote{Refer to Appendix \ref{snwe} for implementation details} and iteratively refines its probing strategies via trial-and-error. Specifically, strategy induction is performed in two stages using distinct simulators and optimization objectives. The first environment focuses on identifying missing or uncertain knowledge areas, inducing \textit{What-to-Ask} strategies that guide the selection of probing targets. The second environment focuses on maximizing information gain while maintaining user engagement, inducing \textit{How-to-Ask} strategies that govern the phrasing and presentation of inquiries.

\begin{algorithm}
\small
\caption{Strategy Induction via Self-play}
\label{alg:code}
\begin{algorithmic}[1] % [1] enables line numbering
\Require $N$: Max simulation sessions, $T$: Max conversation turns
\State \Comment{$W$: \textit{What-to-Ask}, $H$: \textit{How-to-Ask}}
\For{$M \in \{W,H\}$} 
    \State Initialize strategy $S_0^M$
    \State Initialize user simulators $\mathcal{U}^M$

    \For{$k=0$ to $N-1$}
        \State Experience $\mathcal{E} \leftarrow \emptyset$
        \ForAll{$u \in \mathcal{U}^M$}
            \State $c \leftarrow \textsc{SelfPlay}(S_k^M, u, T)$
            \State $\mathcal{E} \leftarrow \mathcal{E} \cup \{c\}$
        \EndFor
        \State $S_{k+1}^M \leftarrow \textsc{Extract\&Refine}(S_k^M,\mathcal{E})$
    \EndFor
\EndFor
\end{algorithmic}
\end{algorithm}

\noindent\textbf{Diverse User Simulation}. Each simulator is instantiated with a unique persona, including \textit{Big-Five Personality Traits}\footnote{Openness, Conscientiousness, Extraversion, Agreeableness, and Neuroticism} \cite{goldberg1992development} and \textit{Decision-Making Styles}\footnote{Directive, Conceptual, Analytical, and Behavioral} \cite{scott1995decision}, to encourage behavioral diversity during interaction. To model varying expertise levels, each simulator is additionally assigned a query and a query-specific expertise label (e.g., \textit{novice}). We further equip the simulator with a query-specific knowledge dependency graph automatically generated by GPT-5, which contains the knowledge required to understand and answer the query. To control the simulator's expertise, we randomly mask different portions of the knowledge dependency graph according to its assigned expertise level. Consequently, novice simulators possess only partial knowledge, whereas expert simulators retain more complete knowledge. This design provides a controllable environment for learning \textit{What-to-Ask} strategies. For \textit{How-to-Ask} strategy induction, the simulator must also exhibit non-cooperative behaviors, as users may refuse to answer poorly phrased or overly demanding questions. Thus, we follow \citet{zhang-etal-2024-strength} and augment the simulators with refusal strategies and corresponding few-shot examples, enabling them to reject inquiries under appropriate circumstances. This forces \ours~to learn questioning strategies that maximize information gain while maintaining user cooperation.

\noindent\textbf{Self-Play Pipeline}. To iteratively optimize probing strategies, we conduct $N$ sessions of offline self-play simulations. We initialize the baseline strategy set $S_0$ using \textit{Gemini 2.5 Pro}. In each $k$-th session, \ours~employs the current strategies ($S_k$) to interact with a user simulator (GPT-4o Mini), aiming to estimate its expertise as described previously. A session terminates upon successful estimation or when the maximum turn limit $T$ is reached. Following each session, we analyze the dialogue logs as experience to induce new strategies, which are then refined to update the set for the next iteration. Refer to Algorithm \ref{alg:code} for details.
\begin{itemize}[leftmargin=*,itemindent=0.05cm, itemsep=-2pt]
    \item \textbf{Strategy Extraction}. Inspired by recent studies in inductive learning, which empowers LLMs to synthesize broader patterns from granular examples \cite{cai2025role, de2025inductive, huang2024araida, wang2025dystil, kim2025principles}, the accumulated self-play experience is distilled into probing strategies that generalize across users and queries. For each self-play stage, we collect simulation dialogue logs together with the probing strategies employed during interaction. We then analyze these trajectories from two complementary perspectives: (1) the accuracy of the final expertise assessment, and (2) the information gain obtained at each probing turn. Using the resulting dialogue trajectory and assessment outcome as context, an LLM is prompted to identify why a particular strategy contributed to successful or failed expertise estimation, or why a probing turn yielded high or low information gain. The extracted insights are subsequently distilled into strategies in a structured form: \textit{Probe [TARGET] to elicit information regarding [ASPECT], because [RATIONALE].}
    \item \textbf{Strategy Refinement}. Since a single session comprises multiple turns, numerous raw strategies are extracted. To prevent these strategies from becoming overly specific to the query or its associated knowledge points, and to avoid redundancy, we employ an LLM-based semantic clustering approach, together with the strategies from previous session. Strategies within the same cluster are then abstracted to derive high-level probing tactics for use in subsequent simulations. 
\end{itemize}

Note that we use the same structured template for both strategy types to control for sentence-structure confounds while allowing their content to differ. Each template specifies the probing target, the expertise-related information sought, and the probe’s utility. What-to-Ask strategies select diagnostic targets, whereas How-to-Ask strategies determine how those targets are elicited conversationally.

\begin{table*}[]
\centering
\resizebox{0.99\textwidth}{!}{%
\begin{tabular}{c|l|cccccc|cccccc}
\toprule
\multicolumn{1}{c|}{\multirow{2}{*}{\textbf{\begin{tabular}[c]{@{}c@{}}LLM \\ Backbone\end{tabular}}}} 
& \multicolumn{1}{c|}{\multirow{2}{*}{\textbf{Method}}} 
& \multicolumn{6}{c|}{\textbf{MMLU-Pro}} 
& \multicolumn{6}{c}{\textbf{ARC-MCAS}} \\ 
\cline{3-14}
\multicolumn{1}{c|}{} 
& \multicolumn{1}{c|}{} 
& \multicolumn{1}{c|}{\textbf{Acc.$\uparrow$}} 
& \multicolumn{1}{c|}{\textbf{Kappa$\uparrow$}} 
& \multicolumn{1}{c|}{\textbf{FA$\uparrow$}} 
& \multicolumn{1}{c|}{\textbf{Comp.$\uparrow$}} 
& \multicolumn{1}{c|}{\textbf{US$\uparrow$}} 
& \multicolumn{1}{c|}{\textbf{Unk.$\downarrow$}} 
& \multicolumn{1}{c|}{\textbf{Acc.$\uparrow$}} 
& \multicolumn{1}{c|}{\textbf{Kappa$\uparrow$}} 
& \multicolumn{1}{c|}{\textbf{FA$\uparrow$}} 
& \multicolumn{1}{c|}{\textbf{Comp.$\uparrow$}} 
& \multicolumn{1}{c|}{\textbf{US$\uparrow$}} 
& \multicolumn{1}{c}{\textbf{Unk.$\downarrow$}} \\ 
\midrule

\multicolumn{14}{c}{\textit{Evaluation with LLM Simulators}}\\
\midrule

\multirow{5}{*}{GPT-4o}
 & IDL (\textit{the best baseline}) 
 & 20.0 & 40.31 & 3.75 & 3.62 & - & 23.79 
 & 20.0 & 15.00 & \textbf{4.59} & 3.42 & - & \underline{4.40} \\

 & Direct Prompt 
 & 16.0 & 41.25 & 3.84 & 3.69 & - & 28.00 
 & \underline{20.0} & 30.64 & \textbf{4.59} & 3.67 & - & 82.80 \\ 
\cmidrule{2-14}

 & \ours 
 & \textbf{77.1} & \textbf{64.90} & \underline{3.85} & \underline{3.83} & \textbf{4.01} & \textbf{0.0} 
 & \textbf{87.4} & \underline{71.28} & \textbf{4.59} & \textbf{3.87} & \textbf{4.18} & \textbf{0.0} \\

 & \ours~w/o What-to-ask 
 & \underline{68.6} & \underline{64.43} & 3.81 & \textbf{3.87} & \underline{3.96} & \textbf{0.0} 
 & \underline{76.8} & 67.26 & 4.45 & \underline{3.86} & 4.08 & \textbf{0.0} \\

 & \ours~w/o How-to-ask 
 & \textbf{77.1} & 64.24 & \textbf{3.93} & 3.86 & 3.97 & \textbf{0.0} 
 & 73.7 & \textbf{71.31} & \underline{4.58} & 3.84 & \underline{4.10} & \textbf{0.0} \\ 

\midrule

\multirow{5}{*}{\begin{tabular}[c]{@{}c@{}}DeepSeek\\ V4-Pro\end{tabular}}
 & IDL (\textit{the best baseline}) 
 & 16.0 & 45.97 & \textbf{4.33} & 3.08 & - & 2.0 
 & 20.0 & 55.70 & 4.49 & 3.62 & - & \textbf{0.0} \\

 & Direct Prompt 
 & 20.0 & \textbf{58.10} & \underline{4.26} & 3.69 & - & \textbf{0.0} 
 & 22.0 & \underline{56.77} & \textbf{4.50} & 3.77 & - & \textbf{0.0} \\ 
\cmidrule{2-14}

 & \ours 
 & \textbf{77.1} & 51.28 & 4.08 & \textbf{3.88} & \textbf{4.45} & \textbf{0.0} 
 & \textbf{78.9} & 52.05 & \textbf{4.50} & \textbf{3.94} & \textbf{4.54} & \textbf{0.0} \\

 & \ours~w/o What-to-ask 
 & \textbf{77.1} & 50.84 & 3.99 & \underline{3.87} & \underline{4.24} & \textbf{0.0} 
 & 68.4 & 56.54 & 4.46 & \underline{3.92} & \underline{4.44} & \textbf{0.0} \\

 & \ours~w/o How-to-ask 
 & \underline{65.7} & \underline{54.80} & 3.91 & 3.81 & 4.17 & \textbf{0.0} 
 & \underline{72.6} & \textbf{57.54} & \textbf{4.50} & 3.90 & 4.33 & \textbf{0.0} \\ 

\midrule

\multicolumn{14}{c}{\textit{Evaluation with Human Participants}}\\ 
\midrule

\multirow{2}{*}{GPT-4o} 
 & IDL 
 & \underline{20} & \underline{40.31} & \underline{3.91} & \underline{3.41} & \underline{3.37} & \underline{23.79} 
 & \underline{20} & \underline{15.00} & \underline{3.73} & \underline{3.47} & \underline{3.55} & \underline{4.40} \\

 & \ours 
 & \textbf{68.7} & \textbf{52.95} & \textbf{4.52} & \textbf{4.62} & \textbf{4.56} & \textbf{0.0} 
 & \textbf{76.7} & \textbf{76.65} & \textbf{4.35} & \textbf{4.74} & \textbf{4.58} & \textbf{0.0} \\ 

\midrule

\multirow{2}{*}{\begin{tabular}[c]{@{}c@{}}DeepSeek\\ V4-Pro\end{tabular}} 
 & IDL 
 & \underline{16} & \underline{45.97} & \underline{3.36} & \underline{3.26} & \underline{3.31} & \underline{2.0} 
 & \underline{20} & \textbf{55.70} & \underline{3.95} & \underline{3.35} & \underline{3.31} & \textbf{0.0} \\

 & \ours 
 & \textbf{67.3} & \textbf{56.98} & \textbf{4.38} & \textbf{4.56} & \textbf{4.43} & \textbf{0.0} 
 & \textbf{56.0} & \underline{38.70} & \textbf{4.64} & \textbf{4.55} & \textbf{4.58} & \textbf{0.0} \\ 

\bottomrule
\end{tabular}%
}
\caption{Main results. Best results are in bold, and second-best results are underlined.}
% \vspace{-3mm}
\label{tab:main_llm}
\end{table*}

%\subsection{Strategy Planner}
%\label{planner}
\subsection{Strategy-guided Inquiry Generation}
\label{inquiry_gen}
Given the populated schema $L_t$ and conversation history $H_t$, \ours~generates the next inquiry in a two-stage manner. First, a \textit{What-to-Ask} strategy is selected based on the current schema and conversation history to determine which aspect of the user's expertise should be further probed. Conditioned on the selected strategy and conversation history, a \textit{How-to-Ask} strategy is then chosen to determine how the inquiry should be phrased. The final inquiry is then generated based on the formulated question and conversation history. After receiving the user's response, \ours~extracts and fills information into the schema based on the user response and conversation history, resulting in an updated schema with filled slots. This process continues until either all schema slots have been populated or the max conversation turn is reached. At that point, \ours~uses the filled schema and conversation history to estimate the user's query-specific expertise level. The estimated level, together with the filled schema and conversation history, is then used to generate a personalized response. Notably, the entire pipeline is implemented through the zero-shot capabilities of LLMs, including strategy selection and expertise estimation. While these components could alternatively be replaced by dedicated trainable modules, the lack of high-quality training data and our exploratory nature of query-specific expertise estimation motivate us to adopt a training-free design in this work. We leave more complex trainable formulations to future research.

\section{Experiments}
\label{exp}

\subsection{Experimental Setup}
We adopt the setup from Section \ref{preE}, with the following key additions to expand our evaluation:

\noindent\textbf{Baselines \& Backbones}. We further involve two LLM backbones (GPT-4o and DeepSeek-V4-Pro) for more comprehensive evaluation. We selected IDL and Direct Prompt for the main comparison. IDL is the strongest baseline overall, with a low rejection rate and stable Kappa scores, while Direct Prompt serves as the most fundamental baseline, representing an agent without explicit user-expertise modeling.

\noindent\textbf{Multi-turn Conversations}. To facilitate the multi-turn conversations required for expertise probing, we employ a two evaluation strategy involving both user simulators and real human participants (cf. Appendix \ref{sim}). For simulator-based evaluation, we adopt the same simulation framework used in \textit{How-to-Ask} strategy induction. To avoid evaluation bias, however, the evaluation simulators are entirely disjoint from those used during strategy induction, with no overlap in personas or refusal strategies.
For human evaluation, we recruit participants to interact directly with \ours~under more realistic settings. Additionally, since expertise levels are defined for each user–query pair rather than for the query alone\footnote{For example, for the same query on quantum mechanics, a computer science PhD student may be a novice, whereas a physics graduate student may be proficient.}, for each query, participants are assigned a target expertise level and instructed to role-play accordingly. To ensure consistency with the assigned query-specific expertise, we provide a knowledge base specifying the concepts expected to be known at that expertise level.

\noindent\textbf{Evaluation Metrics}. Beyond prediction stability, we evaluate Prediction Accuracy (\uline{Acc.}) for query-specific expertise estimation using the ground-truth expertise labels available in our multi-turn conversations. We additionally assess generated responses in terms of factual accuracy (\uline{FA}), comprehensibility (\uline{Comp.}), which evaluates whether responses are appropriately adapted to the user's expertise level; and user satisfaction (\uline{US}), which measures the quality of the overall interaction and the extent to which our probing strategy minimizes user burden. Furthermore, we conduct both human and LLM-based evaluations to assess FA, Comp. and US. Refer to Appendix \ref{human} for details.

\subsection{Main Results}

\noindent\textbf{\ours~consistently establishes superior performance across various LLM backbones and datasets}. As shown in Table \ref{tab:main_llm}, \ours~consistently outperforms the strongest baseline across different LLM backbones and evaluation benchmarks, achieving an average improvement of +288\% in query-level expertise estimation accuracy. Benefiting from more accurate estimation of users' expertise levels, \ours~is able to generate responses that are better tailored to users' knowledge boundaries, substantially improving answer comprehensibility (+4.1\%). Importantly, despite introducing additional probing interactions with users, \ours~maintains factual accuracy comparable to existing baselines, demonstrating that proactive probing does not compromise response correctness. These results suggest that proactively eliciting user expertise provides an effective mechanism for delivering personalized responses.

\noindent\textbf{\ours~demonstrates relatively reliable performance despite the inherent stochasticity of LLM-based probing}. Compared with strong baselines such as IDL and Direct Prompt, \ours~reduces the unknown response rate to nearly zero while maintaining relatively high prediction stability (+42.5\%, on average), indicating that the observed performance gains are not driven by random fluctuations. We also observe that \ours~does not always achieve the highest stability scores. We attribute this to the inherent stochasticity of LLMs in both selecting probing strategies across both the what-to-ask and how-to-ask stages, and estimating user expertise under the predefined schema. Although such variance does not diminish the overall advantage of \ours~over existing methods, it highlights a promising direction for future work: improving system stability through dedicated modules for each stage, especially when supervised training data are available.

\noindent\textbf{Probing strategies in \ours~improve user modeling and help maintain user satisfaction}. Beyond expertise estimation accuracy, \ours~achieves promising user satisfaction scores across different datasets and LLM backbones. Ablation results further show that removing either the \textit{What-to-Ask} or \textit{How-to-Ask} strategy leads to performance degradation, highlighting the importance of both selecting appropriate probing content and generating effective probing interactions. In particular, removing the \textit{How-to-Ask} strategy consistently decreases user satisfaction, suggesting that the manner of questioning substantially affects user experience. Interestingly, removing the \textit{What-to-Ask} strategy also harms satisfaction. We find that, without explicit guidance on probing content, the generated questions become less controllable and occasionally mismatch user expertise levels: overly difficult questions tend to frustrate users, while simpler questions are generally better tolerated.

\noindent\textbf{Human evaluation further validates the effectiveness of \ours}. As shown in Table~\ref{tab:main_llm}, \ours~consistently outperforms IDL across both two backbones under human participants and human evaluation. In particular, \ours~achieves substantially higher expertise estimation accuracy and user satisfaction while reducing the unknown response rate to nearly zero. Meanwhile, the factual accuracy and completeness of generated responses remain competitive, demonstrating that proactive probing can effectively improve personalized interactions without sacrificing response quality. Following \citet{bhagwatkar2025cave}, \citet{zhang-etal-2025-llmtaxo}, and \citet{irawan-etal-2025-towards}, we further examine the consistency between human evaluators and LLM-based evaluators on human interaction data using Gwet's AC2 score \cite{gwet2008computing}. The human--human AC2 reaches an average of 0.72, confirming consistent 
annotation quality across participants. The human--LLM AC2 reaches an 
average of 0.79, indicating strong agreement between human judgments and 
LLM-based evaluation result. Detailed reliability computation and results are provided in 
Appendix~\ref{appendix:iaa}.

\begin{figure}
    \centering
\includegraphics[width=0.95\linewidth]{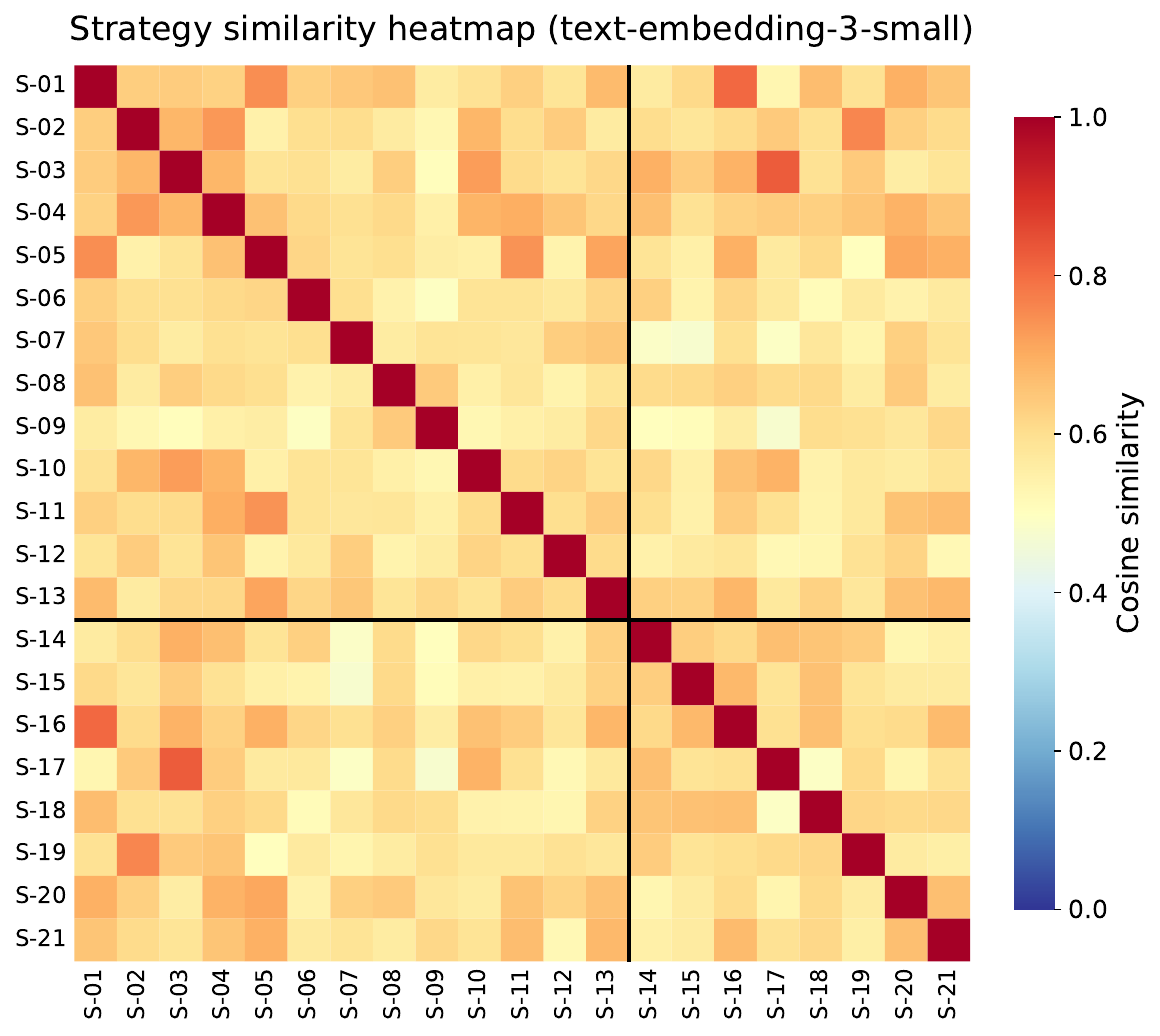}
    \caption{Pairwise similarity among induced strategies (13 What-to-ask and 8 How-to-ask strategies in total)}
    \label{fig:simi}
    % \vspace{-3mm}
\end{figure}

\subsection{In-depth Analysis}
\label{sec:in-depth-analysis}

\noindent\textbf{Induced strategies reveal a functional rather than topical organization}.
We further examine whether the induced What-to-Ask and How-to-Ask strategies occupy distinct regions in embedding space. Using the OpenAI \textit{text-embedding-3-small}, we encode all induced strategies and compute pairwise cosine similarities\footnote{More details are in Appendix~\ref{app:strategy-semantic-analysis}.} (Figure \ref{fig:simi}). The two categories are not semantically separated: the average intra-class similarities are 0.608 for \textit{What-to-Ask} and 0.604 for \textit{How-to-Ask}, while the inter-class similarity is 0.604, with a silhouette score of 0.006. This suggests that the two categories differ primarily in functional role rather than topical content: \textit{What-to-Ask} strategies specify diagnostic targets, whereas \textit{How-to-Ask} strategies operationalize these targets through conversational tactics.

\begin{figure}[t]
\centering
\includegraphics[width=0.98\columnwidth]{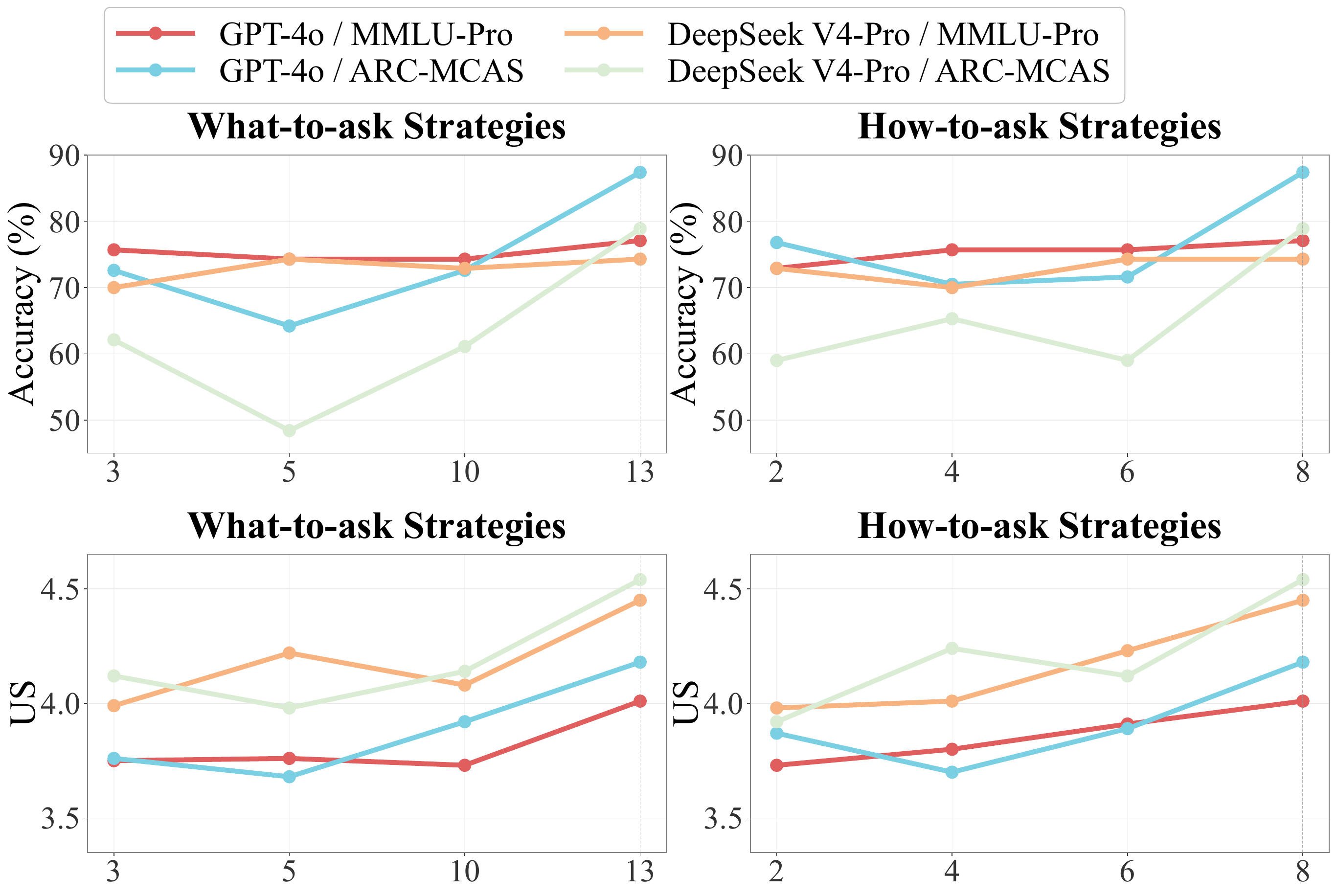}
\caption{Effect of the number of probing strategies on expertise estimation accuracy and user satisfaction.}
% \vspace{-3mm}
\label{fig:strategy_ablation}
\end{figure}

\noindent\textbf{Maintaining sufficient strategies ensures the effectiveness of proactive probing}. We further investigate the impact of the number of probing strategies on both expertise estimation accuracy and user satisfaction in \ours. Specifically, given 13 \textit{What-to-Ask} strategies and 8 \textit{How-to-Ask} strategies, we randomly sample $k$ strategies from one category while keeping the other category complete, and evaluate the resulting performance. As shown in Figure~\ref{fig:strategy_ablation}, increasing the number of strategies generally improves performance, suggesting that richer probing diversity helps the model better capture user expertise and interaction preferences.

\section{Conclusion}
We introduce \ours, a proactive probing method that enables LLM agents to infer query-specific user expertise before generating responses. This is achieved through our LLM-induced two strategy sets. Generally speaking, our findings suggest that effective personalization should not solely depend on passively modeling users from limited observations, but also on actively acquiring missing information through interaction. In this sense, proactive probing serves as a lightweight yet scalable mechanism in real-world interactions. We hope this work can inspire future research on interaction-centric personalization, adaptive information acquisition, and more human-aware LLM agents.

% \clearpage
\section*{Limitations}
As with prior studies on proactive agents by prompting LLMs, the performance of \ours~may be influenced by prompt design. Prompt sensitivity remains an inherent challenge for LLM-based systems, and exploring more robust prompting or training-based alternatives constitutes an important direction for future work. Another limitation of this work is that we do not systematically investigate the impact of self-play hyper-parameters, such as the number of self-play sessions or the maximum conversation turns, on the quality of induced probing strategies. Although we analyze the effect of the number of strategies in our experiments, the strategy induction process itself remains underexplored. Understanding how different self-play configurations influence strategy diversity, quality, and downstream personalization performance is an important direction for future work. Finally, although What-to-Ask and How-to-Ask strategies serve distinct roles, they inevitably exhibit semantic overlap. As an early study of query-specific expertise probing, we leave systematic disentanglement to future work. One promising approach is to condition second-stage extraction on first-stage strategies and explicitly discourage semantic redundancy.

\section*{Ethical Considerations}
Similar to prior work \cite{zhang-etal-2024-clamber, chen-etal-2024-style,10.1145/3269206.3271776, 10.1145/3485447.3512088, chen2023travel}, the primary experiments in this paper rely on LLM-based user simulators for large-scale evaluation. In addition, we conduct supplementary human evaluation experiments with voluntary participants. All participants were informed of the experimental setting and engaged in standard question-answering interactions without sensitive topics or psychologically stressful tasks. Therefore, we believe the study poses minimal ethical risk.

\section*{LLM Usage}
LLMs are used in this work as the backbone models for conversational agents and evaluators in our experiments. In addition, LLMs are also used for language polishing during paper writing. 

\bibliography{custom}

\appendix

\section{Background on User Expertise}
\label{expert}
\subsection{Dreyfus Model}
The Dreyfus Model \cite{dreyfus1980five} describes skill acquisition as a progression through five stages, each characterized by distinct differences across five dimensions: knowledge scope, components, perspective, action, and commitment. Novices possess only explicit factual knowledge, rely entirely on context-free rules, are easily overwhelmed by complex information, reason purely step-by-step, and feel little personal investment in outcomes. Advanced Beginners have slightly broader knowledge but still treat all information as equally relevant and struggle to apply rules to real-world situations. Competent performers mark a turning point: they consciously plan and organize information to filter out noise, and begin to feel personally responsible for their decisions. Proficient users have accumulated rich experiential knowledge that allows them to instantly recognize what matters in a situation without deliberate effort, though they still verify their responses through analytic reasoning. Experts have fully internalized their domain knowledge, rather than recalling facts or following rules, they perceive situations holistically and respond immediately and intuitively.

\subsection{Response Features Required by Experts and Novices}

The substantial differences across Dreyfus levels imply that responses must be tailored accordingly. Novices require short, jargon-free responses that explicitly explain basic concepts in simple language, as they can only process context-free facts and are easily overwhelmed by complex information. Experts, by contrast, expect dense, conclusion-first, telegraphic responses that use domain jargon without explanation and omit all foundational scaffolding.

\section{Experiment Details}
\label{impl}

We use Python 3.10.12, NumPy 1.26.4, and scikit-learn 1.3.2 for implementation and evaluation, including the computation of Fleiss' Kappa~\citep{fleiss1971measuring}. For preliminary experiments, we randomly sample 50 user queries from the MMLU-Pro and ARC-MCAS datasets. For the main experiments, we use the full datasets.

\subsection{Details of Datasets}

\noindent\textbf{MMLU-Pro} \cite{wang2024mmlu} is a challenging multi-task language understanding benchmark spanning 14 academic disciplines. We use 70 questions from the test split for offline self-play strategy induction, and sample 50 questions from the validation split for evaluation to avoid data leakage.

\noindent\textbf{ARC-MCAS} \cite{clark2018think} is a subset of the ARC-Challenge benchmark, consisting of science questions from the Massachusetts Comprehensive Assessment System (MCAS). We sample 95 questions from the test split for evaluation.

\subsection{Implementation Details of Baselines}

\noindent\textbf{Direct Prompt.} The LLM is directly prompted to predict the user's query-specific expertise level given the input query, without any additional reasoning or ensemble steps.

\noindent\textbf{Chain-of-Thought (CoT).} The LLM is prompted to reason step-by-step before predicting the user's expertise level \cite{wei2023chainofthoughtpromptingelicitsreasoning}.

\noindent\textbf{Self-consistency.} Multiple predictions are sampled independently under different random seeds, and the final prediction is determined by majority voting \cite{wang2022self}.

\noindent\textbf{Diverse-Aspect.} The LLM first identifies multiple aspects of the query that are relevant to expertise assessment (e.g., terminology use, reasoning style). For each aspect, an independent expertise judgment is generated. These aspect-level assessments are then integrated into a final cohesive expertise prediction.

\noindent\textbf{IDL.} The LLM is provided with pre-collected dialogue history from the same user as reference. It extracts structured knowledge triples from the history to form a persona constraint, and uses this constraint to assess the user's expertise level on the new query \cite{cheng-etal-2024-dialogues}.

\noindent\textbf{ExpertPrompting.} The LLM is prompted with few-shot examples to assess the user's expertise level directly from the input query, and generates a tailored response accordingly \cite{xu2025expertpromptinginstructinglargelanguage}.

\subsection{Implementation Details of \ours}
As emphasized in prior conversational AI research, agent-initiated probing must account for both users’ willingness to respond and the interaction burden it imposes. Although real users do engage with LLM agents and system-initiated questions in realistic information-seeking scenarios \cite{deng-etal-2023-prompting, 10.1145/3397271.3401160, 3692070.3692401}, demonstrating the practical relevance of this research setting, such engagement should not be assumed to be unconditional or cost-free. Our problem formulation should therefore explicitly model users’ willingness to answer probing questions (i.e., How-to-Ask and What-to-Ask). In particular, our \ours\ aims to provide responses tailored to the user’s query-specific expertise while maintaining user satisfaction. This is achieved by inducing two complementary strategy sets: What-to-Ask selects the most diagnostic expertise evidence, while How-to-Ask phrases the probe in a natural and low-burden way.

To enhance the induction of \textit{What-to-ask strategies}, we further involve a knowledge dependency graph into the \ours's prompt.  
Given a user query $q$ and the model-generated answer\footnote{The specific technique used for answer generation (e.g., retrieval augmentation or asking clarifying questions) is not the primary focus of our work.} $a$, \ours~employs a structured approach to identify and represent the prerequisite knowledge required for comprehending the answer via prompts. During this process, key domain-specific concepts, including terminology entities and actions, are extracted as nodes, while their semantic relationships form the edges. Finally, a knowledge dependency graph $G$ is constructed. Note that each node is additionally assigned a \textit{min-level} attribute indicating the minimum expertise level required to understand that concept; this attribute is used by the user simulator to determine which concepts are known or unknown to a simulated user, but is masked from \ours~during probing to prevent information leakage. In particular, our masking mechanism performs content-based masking according to concept difficulty. Each node in the knowledge dependency graph has a min-level attribute indicating the minimum Dreyfus level required to understand that concept. For a simulator assigned level $l$, we retain nodes whose min-level is no higher than $l$ and mask the remaining nodes. Thus, a Novice retains only Novice-level concepts, an Advanced Beginner retains concepts from the first two levels, and so forth, while an Expert retains all concepts. Since graph size and level distribution vary by query, there is no fixed retention ratio for each level. The min-level attribute is visible only to the user simulator and hidden from \ours\ during probing to prevent information leakage.

\subsection{Implementation Details of User Simulators and Human Participants}
\label{sim}
\subsubsection{User Simulators for Strategy Induction}
\label{snwe}
We detail the expertise Profile for different type of user as follows

\begin{itemize}[leftmargin=*]
    \item The Novice operates with a Knowledge Scope limited to explicit facts and definitions, lacking any deep understanding of "how" things work in practice. They perceive Components strictly as context-free features, relying entirely on rigid rules and abstract principles regardless of the situation. Lacking a functional Perspective, they are often unable to filter information effectively, forcing them to adhere strictly to instructions to avoid becoming overwhelmed. Their Action is purely analytic, characterized by slow, step-by-step reasoning to execute instructions. Consequently, their Commitment remains detached; they view themselves as simply following the rules and feel little personal agency or responsibility for the outcome.
    \item The Advanced Beginner has expanded their Knowledge Scope slightly but still relies heavily on explicit facts. While they begin to recognize some situational Components alongside context-free rules, they lack the experience to distinguish importance. As a result, their Perspective is flawed; they treat all information as equally relevant, which leads them to be easily overwhelmed by the noise of a complex situation. Their Action remains analytic and deliberate as they struggle to apply guidelines to real-world scenarios. Like the Novice, their Commitment is largely detached, as they are focused on surviving the task rather than owning the result.

    \item The Competent performer marks a major shift in Perspective. Faced with a complex situation, they do not get overwhelmed; instead, they consciously make a deliberate plan to organize information and filter out noise. Their Knowledge Scope is now organized enough to handle this structuring. While they perceive both context-free and situational Components, they prioritize them based on a hierarchical goal. Their Action is still analytic—they solve problems by reasoning through their plan step-by-step. However, their Commitment shifts to being involved; because they chose the plan, they feel a sense of personal responsibility and emotional investment in the outcome.

    \item The Proficient user possesses a vast Knowledge Scope of tacit knowledge derived from experience. Their perception of Components is now dominated by situational patterns rather than rules. Crucially, their Perspective has evolved from "making a plan" to "seeing the picture"; they instinctively distinguish important information from noise without conscious effort. Despite this intuitive perception, their Action remains analytic; they effectively diagnose the problem instantly but still rely on step-by-step reasoning to calculate the best specific response. Their Commitment is deeply involved in the perception of the problem, though they may remain analytically detached when executing the solution.

    \item The Expert operates with a deep, tacit Knowledge Scope where knowing "that" has completely transformed into knowing "how." They perceive Components entirely through situational patterns, reading the environment holistically. Their Perspective is immediate and instinctive; important information jumps out at them, and they are never overwhelmed. The defining difference lies in their Action: it shifts from analytic to intuitive. They no longer rely on step-by-step reasoning to decide; the correct response arises immediately and fluidly. Their Commitment is fully involved, acting with a level of agency where the person and the task are effectively merged.

\end{itemize}

\textit{What-to-Ask} vs.\ \textit{How-to-Ask} Simulator Design:

\begin{itemize}[leftmargin=*]
    \item While both simulators instantiate users according to the five Dreyfus profiles described above, they differ in design. The \textit{What-to-Ask} simulator injects only the expertise-level profile and knowledge boundaries (known/unknown concepts), and the simulator's sole task is to generate a response consistent with the cognitive style of that level; responses are output as plain text with no explicit decision process, yielding five fixed behavioral distributions. The \textit{How-to-Ask} simulator builds on this foundation by overlaying a personality trait drawn from the Big Five model \cite{goldberg1992development} and a decision-making style drawn from four archetypes \cite{scott1995decision}, allowing users at the same expertise level to exhibit markedly different communicative styles. It further introduces two additional mechanisms: a level-specific refusal pattern that characterizes how users at each expertise level typically decline to answer, and an explicit answer/refuse decision step that requires the simulator to weigh reasons for answering against reasons for refusing before producing a structured output containing both a decision and a response. The resulting training dialogues cover uncooperative behaviors including deflection, minimal engagement, and outright refusal, forcing the how-to-ask strategies to learn how to adapt their phrasing when users are not forthcoming.
\end{itemize}

\subsubsection{User Simulators for Experimental Evaluation}
\label{sim_main}

For the main experiments, we employ GPT-4o Mini as the user simulator. Each simulator instance is initialized with a pre-defined expertise profile comprising: (1) an expertise level drawn from the five Dreyfus levels; (2) a persona description including personality trait based on the Big Five model \cite{goldberg1992development} and decision-making style \cite{scott1995decision}; (3) a set of known and unknown concepts derived from the knowledge dependency graph. To better simulate realistic user behavior, each expertise level is assigned a corresponding refusal pattern that allows the simulator to autonomously decide whether to answer or refuse a given inquiry. The simulator strictly respects its knowledge boundaries by never displaying knowledge of unknown concepts, enabling diverse and realistic simulation of users across different expertise levels.

\subsection{Implementation Details of Evaluation Metrics}
\label{metrics}
\noindent\textbf{Prediction Stability (Kappa).} We measure prediction stability using Fleiss' Kappa~\citep{fleiss1971measuring}, defined as:
\begin{equation*}
\kappa = \frac{\bar{P} - \bar{P}_e}{1 - \bar{P}_e},
\end{equation*}
where $\bar{P}$ denotes the mean observed agreement across all queries, and $\bar{P}_e$ denotes the expected agreement under chance. To quantify cross-run prediction consistency, we treat predictions produced under different random seeds as independent raters. The resulting Kappa score reflects how stable the method is across repeated runs.

\noindent\textbf{Factual Accuracy (FA).} We employ GPT-5 as a third-party evaluator to assess the factual accuracy of the tailored responses. The evaluation covers four dimensions: core answer correctness, detail correctness, no factual error, and no misleading, each scored on a 1--5 scale. The final FA score is the average across all four dimensions.

\noindent\textbf{Comprehension (Comp.).} Comprehension is evaluated by the user simulator (GPT-4o Mini) from the perspective of the simulated user, initialized with a pre-defined expertise profile including known and unknown concepts. The simulator assesses four dimensions: clarity, level match, completeness, and accessibility, each scored on a 1--5 scale. The final Comp. score is the average across all four dimensions.

\noindent\textbf{User Satisfaction (US).} We employ GPT-5 as a third-party evaluator to assess user satisfaction based on the full conversation history. The evaluation covers five dimensions: comfort, feeling understood, willingness to engage, conversation naturalness, and overall satisfaction, each scored on a 1--5 scale. The final US score is the average across all five dimensions.
\subsection{Implementation Details of Human Evaluation}
\label{human}

\subsubsection{Role-playing Protocol}
\label{human_protocol}

We recruited ten graduate students specializing in computer science and artificial intelligence to participate in the human evaluation. All participants voluntarily took part in the study.

Each participant role-played users at five expertise levels defined by the Dreyfus Model: Novice, Advanced Beginner, Competent, Proficient, and Expert.
The expertise profiles used for role-playing follow the same Dreyfus-based descriptions as those used in our user simulators for strategy induction, as detailed in Appendix~\ref{snwe}.
Each participant completed 10 conversations per level, yielding 50 conversations per participant. Experiments were conducted on two datasets, MMLU-Pro and ARC-MCAS, and two LLM backbones, GPT-4o and DeepSeek-V4-Pro.

To ensure consistent role-playing behavior across participants, we provided participants with the expertise-level guide described in Appendix~\ref{snwe}, together with representative example responses.
This design allowed participants to focus on controlling the reasoning style, uncertainty, and response granularity associated with each target level.

After each conversation, participants evaluated the system's tailored response on three dimensions using 5-point Likert scales.
Factual Accuracy (FA, 4 items) measures factual consistency between the system response and a provided reference answer, independent of expertise-level adaptation.
Comprehensibility (Comp, 4 items) assesses whether the response is clear and understandable for the role-played user level.
User Satisfaction (US, 5 items) measures whether the response is appropriate, helpful, and well aligned with the participant's assigned expertise level.

\subsection{Implementation Details of Experiment}
\paragraph{Semantic analysis of induced strategies.}
\label{app:strategy-semantic-analysis}

We provide additional details for the semantic analysis in Section~\ref{sec:in-depth-analysis}. We encode all 21 induced strategies using an OpenAI-compatible embedding endpoint based on text-embedding-3-small, obtaining a 1536-dimensional embedding for each strategy. The strategy set contains 13 \textit{What-to-Ask} strategies and 8 \textit{How-to-Ask} strategies. We compute pairwise cosine similarities and report intra-class similarity, inter-class similarity, the diversity gap, and the silhouette score under the \textit{What-to-Ask}/\textit{How-to-Ask} labeling.

The diversity gap is defined as:
\[
\Delta =
\frac{\mathrm{Sim}_{\textit{What-to-Ask}} + \mathrm{Sim}_{\textit{How-to-Ask}}}{2}
-
\mathrm{Sim}_{\mathrm{inter}},
\]
where \(\mathrm{Sim}_{\textit{What-to-Ask}}\) and \(\mathrm{Sim}_{\textit{How-to-Ask}}\) are the average within-category cosine similarities excluding diagonal self-similarities, and \(\mathrm{Sim}_{\mathrm{inter}}\) is the average cross-category cosine similarity.

\begin{table}[t]
\centering
% \small
\begin{tabular}{lc}
\toprule
Metric & Value \\
\midrule
\textit{What-to-Ask} intra-class similarity & 0.608 \\
\textit{How-to-Ask} intra-class similarity & 0.604 \\
Inter-class similarity & 0.604 \\
Diversity gap \(\Delta\) & 0.003 \\
Silhouette score & 0.006 \\
\bottomrule
\end{tabular}
\caption{Semantic similarity statistics for induced strategies.}
\label{tab:strategy-semantic-stats}
\end{table}

The closest cross-category pairs are S-03/S-17 (0.825), S-01/S-16 (0.807), and S-02/S-19 (0.759), corresponding to conceptual distinction, rationale probing, and concrete application, respectively. These nearest-neighbor pairs support the interpretation that \textit{How-to-Ask} strategies specify \textit{How-to-Ask} questions that elicit evidence for diagnostic targets defined by \textit{What-to-Ask} strategies.

\subsection{Human Evaluation Reliability}
\label{appendix:iaa}

\paragraph{Setup.}
To assess the reliability of the human evaluation described in
Section~\ref{human_protocol}, we compute inter-rater agreement from two
perspectives. First, we measure \textit{human--human} agreement among the human participants to verify that the evaluation criteria were applied
consistently. Second, we measure \textit{human--LLM} agreement between the
human ratings and the LLM-based evaluator to examine whether the automatic
evaluator aligns with human judgment. Reliability is computed separately for
the three post-conversation evaluation dimensions: factual accuracy (FA),
comprehensibility (Comp.), and user satisfaction (US).

\paragraph{Metric and computation.}
We use ordinal-weighted \textbf{Gwet's AC2} \cite{gwet2008computing} as the
reliability metric. AC2 is a chance-corrected agreement coefficient that
compares the observed agreement among raters with the agreement expected by
chance:
\[
\mathrm{AC2} = \frac{A_o - A_e}{1 - A_e},
\]
where $A_o$ denotes the observed weighted agreement and $A_e$ denotes the
chance-expected agreement estimated under Gwet's occasional-guessing model.
Since our evaluation uses 1--5 Likert-scale scores, we apply ordinal weights
so that larger score differences incur larger disagreement penalties.

For human--human reliability, AC2 is computed over all independent human
ratings for each evaluated session. For human--LLM reliability, we compare the
LLM-based evaluator's score with the corresponding aggregated human score for
each session. Since each evaluation dimension consists of multiple Likert-scale
sub-items, we first aggregate the sub-items within each dimension and then
compute AC2 on the resulting dimension-level scores.

\paragraph{Results.}
As shown in Table~\ref{tab:iaa}, human--human AC2 ranges from 0.52 to 0.87
($\mu=0.72$), indicating moderate-to-strong agreement among participants.
Human--LLM AC2 ranges from 0.74 to 0.85 ($\mu=0.79$), which is comparable
to human--human agreement and suggests that the LLM-based evaluator is well aligned with human judgment. Furthermore, Table \ref{tab:passing_us_comparison} reports bootstrap 95\% confidence intervals as error bars for the US scores in Table \ref{tab:main_llm}. We conduct participant-level paired analysis: for each participant, we first aggregate the US scores under IDL and \ours\, then compute paired differences. We use an exact paired sign-flip test, with Holm correction across the four backbone–dataset settings. Across all four settings, \ours\ achieves significantly higher US than IDL, with paired differences ranging from 1.028 to 1.280; all 95\% CIs are strictly positive, and the Holm-corrected p-values are 0.007812. This confirms that \ours\'s user satisfaction is significantly higher than IDL under participant-level paired analysis. 

\begin{table*}[t]
\centering
% \small
\begin{tabular}{lccc}
\toprule
\textbf{Setting} 
& \textbf{\ours\ US (95\% CI)} 
& \textbf{Difference (95\% CI)} 
& \textbf{Holm $p$} \\
\midrule
GPT-4o / MMLU-Pro
& 4.580 [4.540, 4.621]
& 1.224 [1.073, 1.360]
& 0.007812 \\

GPT-4o / ARC-MCAS
& 4.583 [4.573, 4.594]
& 1.028 [0.999, 1.055]
& 0.007812 \\

DeepSeekV4-Pro / MMLU-Pro
& 4.423 [4.408, 4.436]
& 1.146 [1.056, 1.234]
& 0.007812 \\

DeepSeekV4-Pro / ARC-MCAS
& 4.568 [4.558, 4.580]
& 1.280 [1.240, 1.321]
& 0.007812 \\
\bottomrule
\end{tabular}
\caption{Error Bars and Statistical Significance in Human Evaluation.}
\label{tab:passing_us_comparison}
\end{table*}

\begin{table}[t]
\centering
\small
\setlength{\tabcolsep}{4pt}
\begin{tabular}{llccc}
\toprule
\multirow{2}{*}{\textbf{Dataset}} 
& \multirow{2}{*}{\textbf{Backbone}}
& \multicolumn{3}{c}{\textbf{Gwet's AC2} $\uparrow$} \\
\cmidrule(lr){3-5}
& & \textbf{FA} & \textbf{Comp.} & \textbf{US} \\
\midrule
\multicolumn{5}{l}{\textit{Human--Human}} \\
\midrule
\multirow{2}{*}{ARC-MCAS}  
  & DeepSeek-V4-Pro & 0.7681 & 0.6996 & 0.6641 \\
  & GPT-4o          & 0.6949 & 0.8488 & 0.7614 \\
\addlinespace[2pt]
\multirow{2}{*}{MMLU-Pro} 
  & DeepSeek-V4-Pro & 0.8698 & 0.8121 & 0.7888 \\
  & GPT-4o          & 0.6213 & 0.5153 & 0.6477 \\
\midrule
\multicolumn{2}{l}{\textit{Average}} & 0.7385 & 0.7190 & 0.7155 \\
\midrule
\multicolumn{5}{l}{\textit{Human--LLM}} \\
\midrule
\multirow{2}{*}{ARC-MCAS}  
  & DeepSeek-V4-Pro & 0.7436 & 0.7563 & 0.8096 \\
  & GPT-4o          & 0.7351 & 0.7954 & 0.7887 \\
\addlinespace[2pt]
\multirow{2}{*}{MMLU-Pro} 
  & DeepSeek-V4-Pro & 0.8277 & 0.7479 & 0.7645 \\
  & GPT-4o          & 0.7766 & 0.8489 & 0.7419 \\
\midrule
\multicolumn{2}{l}{\textit{Average}} & 0.7708 & 0.7871 & 0.7762 \\
\bottomrule
\end{tabular}
\caption{Inter-rater agreement (Gwet's AC2) for human--human and human--LLM evaluation across datasets and backbones. FA: factual accuracy; Comp.: comprehensibility; US: user satisfaction.}
\label{tab:iaa}
\end{table}

\subsection{Additional Analysis}
For better understanding, we encode the five Dreyfus levels from 0 to 4 and compute mean absolute error between predicted and ground-truth levels. On MMLU-Pro and ARC-MCAS, the MAEs are 0.343 and 0.147 for GPT-4o, and 0.243 and 0.211 for DeepSeekV4-Pro, respectively. When pooling both datasets, the MAEs are 0.230 for GPT-4o and 0.224 for DeepSeekV4-Pro. Overall, 96.97\% of predictions are within one level of the ground truth. We will report these ordinal metrics in the revised manuscript.

We also include the corresponding 5×5 confusion matrices in Table \ref{tab:confusion_matrices}, where rows denote ground-truth Dreyfus levels and columns denote predicted levels in the order Novice, Advanced Beginner, Competent, Proficient, Expert.

\begin{table*}[t]
\centering
\caption{Confusion matrices across different LLM backbones and datasets.
Rows denote ground-truth Dreyfus levels and columns denote predicted levels in the order Novice, Advanced Beginner, Competent, Proficient, Expert.}
\label{tab:confusion_matrices}
\setlength{\tabcolsep}{3.5pt}
\renewcommand{\arraystretch}{1.08}

\resizebox{\textwidth}{!}{
\begin{tabular}{
lrrrrr @{\hspace{1.2em}}
lrrrrr @{\hspace{1.2em}}
lrrrrr @{\hspace{1.2em}}
lrrrrr
}
\toprule
\multicolumn{6}{c}{\textbf{GPT-4o / MMLU-Pro}}
&
\multicolumn{6}{c}{\textbf{GPT-4o / ARC-MCAS}}
&
\multicolumn{6}{c}{\textbf{DeepSeekV4-Pro / MMLU-Pro}}
&
\multicolumn{6}{c}{\textbf{DeepSeekV4-Pro / ARC-MCAS}}
\\
\cmidrule(lr){1-6}
\cmidrule(lr){7-12}
\cmidrule(lr){13-18}
\cmidrule(lr){19-24}

& N & AB & C & P & E
& & N & AB & C & P & E
& & N & AB & C & P & E
& & N & AB & C & P & E \\
\midrule

N
& \textbf{14} & 0 & 0 & 0 & 0
& N
& \textbf{18} & 1 & 0 & 0 & 0
& N
& \textbf{14} & 0 & 0 & 0 & 0
& N
& \textbf{18} & 1 & 0 & 0 & 0 \\

AB
& 0 & \textbf{12} & 1 & 1 & 0
& AB
& 2 & \textbf{17} & 0 & 0 & 0
& AB
& 6 & \textbf{8} & 0 & 0 & 0
& AB
& 4 & \textbf{15} & 0 & 0 & 0 \\

C
& 0 & 1 & \textbf{13} & 0 & 0
& C
& 0 & 1 & \textbf{18} & 0 & 0
& C
& 0 & 2 & \textbf{9} & 3 & 0
& C
& 0 & 3 & \textbf{13} & 3 & 0 \\

P
& 0 & 0 & 0 & \textbf{14} & 0
& P
& 0 & 0 & 0 & \textbf{19} & 0
& P
& 0 & 1 & 0 & \textbf{13} & 0
& P
& 0 & 0 & 1 & \textbf{16} & 2 \\

E
& 0 & 1 & 5 & 7 & \textbf{1}
& E
& 0 & 0 & 2 & 6 & \textbf{11}
& E
& 0 & 0 & 0 & 4 & \textbf{10}
& E
& 0 & 0 & 0 & 6 & \textbf{13} \\

\bottomrule
\end{tabular}
}
\end{table*}

\section{Case Study}

\noindent
We provide a complete case study illustrating the full \ours\ pipeline:
initial query $\rightarrow$ probing questions $\rightarrow$ user responses
$\rightarrow$ schema filling $\rightarrow$ expertise estimation
$\rightarrow$ tailored response.

\begin{tcolorbox}[querybox]
\textbf{Query:} “Why can large-aperture telescopes see fainter objects?”
\end{tcolorbox}

\begin{tcolorbox}[turnbox,title={Turn 1}]
\begin{itemize}[
  leftmargin=5mm,
  itemsep=2pt,
  topsep=1pt,
  parsep=0pt
]
  \item What-to-Ask: S-03 (The strategy descriptions are provided at the end)
  \item How-to-Ask: S-14
  \item Inquiry: “This touches on a bit of optics — no worries if it’s unfamiliar! Do you have a sense of what light-gathering power actually refers to, or how it differs from magnification?”
  \item User response: “I know a bigger aperture lets you see more, but I always thought that was mainly about zooming in, not really about brightness.”
  \item Schema update: knowledge\_scope $\rightarrow$ explicit; perspective $\rightarrow$ overwhelmed
\end{itemize}
\end{tcolorbox}

\begin{tcolorbox}[turnbox,title={Turn 2}]
\begin{itemize}[
  leftmargin=5mm,
  itemsep=2pt,
  topsep=1pt,
  parsep=0pt
]
  \item What-to-Ask: S-02
  \item How-to-Ask: S-18
  \item Inquiry: “Here’s a comparison: if you wanted to collect more rainwater, what would you do? How might that relate to how a telescope collects light?”
  \item User response: “Um, maybe a wider opening catches more rain? So maybe a bigger telescope catches more light too, I guess.”
  \item Schema update: components $\rightarrow$ context\_free; action $\rightarrow$ intuitive; commitment $\rightarrow$ detached
\end{itemize}
\end{tcolorbox}

\begin{tcolorbox}[resultbox]
\textbf{Termination and Estimation} All five schema slots are populated within 2 turns. \ours\ estimates the user as Advanced Beginner.
\end{tcolorbox}

\begin{tcolorbox}[resultbox]
\textbf{Final Tailored Answer} You’re on the right track! A telescope’s light-gathering power depends on the area of its aperture, not just the width — and area scales with the square of the diameter. So doubling the aperture roughly quadruples the light collected, letting it pick up much fainter objects.
\end{tcolorbox}

\begin{tcolorbox}[
  strategybox,
  title={Strategy IDs used in the case example:}
]
\begin{itemize}[
  leftmargin=5mm,
  itemsep=4pt,
  topsep=1pt,
  parsep=0pt
]
  \item S-02 (What-to-Ask): Probe its application in a different context (e.g., a practical scenario, a general rule, or a self-generated example) to elicit information regarding action, because this tests the ability to bridge declarative and procedural knowledge, distinguishing academic recall or guided analysis from a robust, applicable, and transferable understanding.

  \item S-03 (What-to-Ask): Probe the key distinction from a similar concept, the precise mechanism behind their intuitive language, or the nuances omitted in their simplification to elicit information regarding knowledge\_scope, because this moves beyond simple recall to test the user’s understanding of conceptual boundaries, surfacing foundational misunderstandings and revealing the true depth of their knowledge.

  \item S-14 (How-to-Ask): Probe a single, simplified component of the difficult concept — first acknowledging the difficulty to normalize it, then asking a more focused question — to elicit information regarding knowledge\_scope, because it reduces cognitive load and anxiety, preventing disengagement while establishing a baseline of knowledge from which to build.

  \item S-18 (How-to-Ask): Probe their interpretation of or first step within a brief, concrete situation provided as a shared reference point to elicit information regarding perspective, because it lowers initial cognitive load and provides a common anchor, effectively grounding the conversation for all expertise levels and preventing novices from stalling.
\end{itemize}
\end{tcolorbox}

To understand why IDL struggles to estimate user expertise, we compare IDL with \ours{} on two representative cases with the same gold level but different predictions.
As shown in Table~\ref{tab:case_comparison}, IDL makes opposite errors: it underestimates a \textit{Competent} user in computer science when the history resembles closed-form QA, while overestimating a \textit{Competent} user in chemistry when the history contains advanced terms such as group-14 hydrides and EPR spectroscopy.
This reveals a shared failure mechanism: IDL mistakes question-level signals for user-level understanding.

\begin{table*}[t]
\centering
\footnotesize
\setlength{\tabcolsep}{4pt}
\renewcommand{\arraystretch}{1.22}

\newcolumntype{C}[1]{>{\centering\arraybackslash}m{#1}}
\newcolumntype{Y}{>{\centering\arraybackslash}X}

\begin{tabularx}{\textwidth}{
@{}
C{1.0cm}
C{2.0cm}
C{1.5cm}
C{2.3cm}
C{1.8cm}
Y
Y
@{}
}
\toprule
\textbf{Case} 
& \textbf{Domain} 
& \textbf{Expertise} 
& \textbf{IDL Pred.} 
& \textbf{\ours{} Pred.} 
& \textbf{IDL Error Pattern} 
& \textbf{\ours{} Evidence} \\
\midrule

Case 1 
& Computer Science 
& Competent 
& Advanced Beginner 
& Competent
& Underestimates the user because closed-form QA history appears shallow.
& Probes concept boundaries, mechanism understanding, and applied judgment. \\

\addlinespace[2pt]

Case 2 
& Chemistry 
& Competent 
& Proficient 
& Competent
& Overestimates the user because advanced terminology appears expert-like.
& Probes mechanism understanding, uncertainty, and guided reasoning. \\

\bottomrule
\end{tabularx}

\caption{Case-level comparison between IDL and \ours{}. }
\label{tab:case_comparison}
\end{table*}

These cases suggest that IDL fails because it relies on weak proxy signals rather than direct evidence of user reasoning ability.
It observes historical question difficulty, domain consistency, and interaction format, but cannot observe how the user reasons through a concept.
As a result, IDL infers expertise from what topics the user has asked about, rather than what the user actually understands.

In contrast, \ours{} actively constructs diagnostic evidence through follow-up questions.
In the computer science case, \ours{} probes concept boundaries, encryption mechanisms, and zero-day disclosure trade-offs.
In the chemistry case, \ours{} asks about M--H bond strength and further narrows the discussion to orbital overlap, revealing whether the user understands the underlying mechanism or only recognizes the general trend.

Overall, the case study shows that accurate user-level perception requires observing how users reason, not merely what topics they ask about.
IDL performs proxy-based inference from historical questions, whereas \ours{} performs reasoning-based diagnosis through adaptive questioning.
Across cases, \ours{} follows a consistent pattern: probing concept boundaries, mechanisms, uncertainty, and applied judgment.
Thus, when historical signals are misleading, \ours{} can still recover the correct user level.
In short, IDL follows ``historical question difficulty $\rightarrow$ proxy-based expertise inference'', whereas \ours{} follows ``adaptive follow-up dialogue $\rightarrow$ reasoning-based expertise diagnosis''.

\section{Induced Strategies}

\subsection{What-to-Ask Strategies}

\begin{description}

\item[S-01]
Probe their step-by-step plan or the underlying rationale for a specific choice to elicit information regarding \textit{perspective}, because this externalizes the user's mental model, revealing their procedural thinking, priorities, and whether their actions are based on rote memorization or goal-oriented understanding.

\item[S-02]
Probe its application in a different context (e.g., a practical scenario, a general rule, or a self-generated example) to elicit information regarding \textit{action}, because this tests the ability to bridge declarative and procedural knowledge, distinguishing academic recall or guided analysis from a robust, applicable, and transferable understanding.

\item[S-03]
Probe the key distinction from a similar concept, the precise mechanism behind their intuitive language, or the nuances omitted in their simplification to elicit information regarding \textit{knowledge\_scope}, because this moves beyond simple recall to test the user's understanding of conceptual boundaries, surfacing foundational misunderstandings and revealing the true depth of their knowledge.

\item[S-04]
Probe a boundary case, exception, ambiguous scenario, or common pitfall that challenges its standard application to elicit information regarding \textit{perspective}, because this differentiates rigid, rule-based knowledge from a nuanced, adaptive understanding of a concept's limitations and failure modes in complex, real-world situations.

\item[S-05]
Probe the relative importance of factors, alternative approaches, trade-offs, or the limitations and critiques of their chosen approach to elicit information regarding \textit{perspective}, because this reveals whether the user has a holistic, prioritized mental model that considers the problem space broadly, including counterarguments and limitations, rather than a linear or one-sided view.

\item[S-06]
Probe the precise, mechanistic interaction between components or other interacting systemic factors to elicit information regarding \textit{components}, because this distinguishes static, list-based knowledge from a dynamic, systems-level understanding by forcing the user to trace causal links and consider the system holistically.

\item[S-07]
Probe the informal signals, `gut feelings,' or intuitive patterns they use to guide analysis or form hypotheses to elicit information regarding \textit{action}, because this differentiates a competent user's deliberate process from an expert's intuitive pattern-matching, revealing the shift from analytical to holistic decision-making.

\item[S-08]
Probe the specific source of their hesitation or their method for reconciling the inconsistency to elicit information regarding \textit{commitment}, because this forces the user to articulate the precise boundaries of their own knowledge, revealing how their knowledge is structured and whether they can resolve internal conflicts.

\item[S-09]
Probe their personal experiences, professional stance, or the most difficult `real-world' aspect of the issue to elicit information regarding \textit{commitment}, because an expert's deep experience is often linked to a strong sense of personal or professional investment, and eliciting this can help differentiate a detached analyst from an involved expert.

\item[S-10]
Probe the underlying first principles, policy reasons, or historical context for that rule to elicit information regarding \textit{knowledge\_scope}, because this tests for tacit understanding of a rule's origin and purpose (`the why') rather than just explicit knowledge of its content (`the what'), a key differentiator between Competent practitioners and Proficient or Expert thinkers.

\item[S-11]
Probe the limitations, critiques, or weaknesses of that same concept or model to elicit information regarding \textit{perspective}, because a Competent user can accurately apply a model, but an Expert understands its boundaries and when it breaks down. This probe forces a shift from a deliberate explanation of the model's parts to a holistic evaluation of the model itself.

\item[S-12]
Probe a counter-intuitive scenario where that causal link is broken or reversed to elicit information regarding \textit{action}, because analytic reasoning follows predictable causal chains, which is characteristic of Competence. Asking for exceptions or paradoxes tests for an intuitive grasp of the topic, which allows for non-linear thinking and is a hallmark of an Expert.

\item[S-13]
Probe a request for synthesis and prioritization across the discussed topics (e.g., `What is the single most overlooked factor?') to elicit information regarding \textit{perspective}, because such signals suggest a user may have a deeper, more integrated understanding than they have explicitly stated. Asking them to synthesize and prioritize forces them to move beyond explaining individual components and demonstrate a holistic view of the entire system, revealing Proficient or Expert-level thinking.

\end{description}

\subsection{How-to-Ask Strategies}

\begin{description}

\item[S-14]
Probe a single, simplified component of the difficult concept --- first acknowledging the difficulty to normalize it, then asking a more focused question --- to elicit information regarding \textit{knowledge\_scope}, because it reduces cognitive load and anxiety, preventing disengagement while establishing a baseline of knowledge from which to build.

\item[S-15]
Probe the consequence or increasing complexity of their correct foundational answer --- building on it affirmingly to incrementally raise the scope (e.g., `Exactly. Building on that, what is the consequence of...?') --- to elicit information regarding \textit{perspective}, because it creates a logical and encouraging conversational flow, confirming their base knowledge while smoothly gauging the depth of their understanding.

\item[S-16]
Probe the `why' behind their `what' --- the underlying principle or reasoning they have not yet justified (e.g., `Can you walk me through your thinking?' or `What's the underlying principle for that step?') --- to elicit information regarding \textit{perspective}, because it elicits the user's mental model, distinguishing deep, principle-based understanding from rote memorization.

\item[S-17]
Probe the precise distinction between two closely related concepts --- whether the user has defined one correctly or only partially --- (e.g., `What's the primary difference between X and Y?') to elicit information regarding \textit{knowledge\_scope}, because it tests the precision and boundaries of their understanding, revealing whether their knowledge is superficial or nuanced and interconnected.

\item[S-18]
Probe their interpretation of or first step within a brief, concrete situation provided as a shared reference point to elicit information regarding \textit{perspective}, because it lowers initial cognitive load and provides a common anchor, effectively grounding the conversation for all expertise levels and preventing novices from stalling.

\item[S-19]
Probe a concrete, self-generated example from their own experience that grounds the abstract concept they described (e.g., `Can you give me a specific example of when you've seen that happen?') to elicit information regarding \textit{components}, because it tests for active recall and grounds abstract knowledge in practical application, differentiating between theoretical knowledge and applied expertise.

\item[S-20]
Probe how their approach would shift when a trade-off, conflicting goal, or ambiguous constraint is introduced into the scenario (e.g., `How would you prioritize if you had limited resources?' or `What if constraint X were introduced?') to elicit information regarding \textit{action}, because it tests the robustness and flexibility of a user's mental model, revealing their ability to adapt and make judgments under pressure --- a hallmark of true expertise.

\item[S-21]
Probe how a different stakeholder would view or evaluate the same concept or decision (e.g., `How would the finance team view this decision differently?') to elicit information regarding \textit{perspective}, because it probes for a systems-level understanding and empathy, which are hallmarks of higher proficiency, moving beyond self-contained knowledge.

\end{description}

\section{Prompts}
\begin{center}
\begin{tabular}{p{0.92\linewidth}}
\toprule
\textit{\textbf{The knowledge graph construction prompt}} \\
\midrule
\textbf{Input:} \textless{}Query\textgreater{}, \textless{}Answer\textgreater{} \\[4pt]
\textbf{Output 1:} \textless{}Knowledge Dependency Graph with min-level\textgreater{} (for user simulator) \\[4pt]
\textbf{Output 2:} \textless{}Knowledge Dependency Graph without min-level\textgreater{} (for probing) \\
\bottomrule
\end{tabular}
\captionof{table}{The KG construction. The \textit{min-level} attribute is masked from the probing agent to prevent information leakage.}
\label{tab:prompt_kg}
\end{center}
\vspace{-8pt}

\begin{center}
\begin{tabular}{p{0.92\linewidth}}
\toprule
\textit{\textbf{The \textit{What-to-Ask} strategy selection prompt}} \\
\midrule
Your task is to decide what information to ask about, using the \textit{What-to-Ask} strategy set. \\[4pt]
\textbf{Inputs:} \\
\quad \textless{}Conversation History\textgreater{} \\
\quad \textless{}Populated Schema\textgreater{} \\
\quad \textless{}\textit{What-to-Ask} Strategy Set\textgreater{} \\[4pt]
\textbf{Output:} \textless{}Information to Ask\textgreater{} \\
\bottomrule
\end{tabular}
\captionof{table}{The \textit{What-to-Ask} prompt for deciding what information to ask about.}
\label{tab:prompt_wta}
\end{center}
\vspace{-8pt}

\begin{center}
\begin{tabular}{p{0.92\linewidth}}
\toprule
\textit{\textbf{The \textit{How-to-Ask} strategy selection prompt}} \\
\midrule
Your task is to decide how to ask that question, using the \textit{How-to-Ask} strategy set. \\[4pt]
\textbf{Inputs:} \\
\quad \textless{}Information to Ask\textgreater{} \\
\quad \textless{}Conversation History\textgreater{} \\
\quad \textless{}\textit{How-to-Ask} Strategy Set\textgreater{} \\[4pt]
\textbf{Output:} \textless{}Formulated Question\textgreater{} \\
\bottomrule
\end{tabular}
\captionof{table}{The \textit{How-to-Ask} prompt for deciding how to ask that question.}
\label{tab:prompt_hta}
\end{center}
\vspace{-8pt}

\begin{center}
\begin{tabular}{p{0.92\linewidth}}
\toprule
\textit{\textbf{The inquiry generation prompt}} \\
\midrule
Your task is to generate the inquiry to ask the user. \\[4pt]
\textbf{Inputs:} \\
\quad \textless{}Formulated Question\textgreater{} \\
\quad \textless{}Conversation History\textgreater{} \\[4pt]
\textbf{Output:} \textless{}Inquiry\textgreater{} \\
\bottomrule
\end{tabular}
\captionof{table}{The inquiry generation prompt.}
\label{tab:prompt_stage3}
\end{center}
\vspace{-8pt}

\begin{center}
\begin{tabular}{p{0.92\linewidth}}
\toprule
\textit{\textbf{The user simulator prompt}} \\
\midrule
Your task is to roleplay as a real person and respond to the inquiry, strictly following the persona, expertise level, and knowledge boundaries specified in the inputs below. \\[4pt]
\textbf{Inputs:} \\
\quad \textless{}Persona Description\textgreater{} \\
\quad \textless{}Dreyfus Level\textgreater{}, \textless{}Known Concepts\textgreater{}, \textless{}Unknown Concepts\textgreater{} \\
\quad \textless{}Refusal Strategies\textgreater{} \\
\quad \textless{}User Query\textgreater{}, \textless{}Conversation History\textgreater{}, \textless{}Inquiry\textgreater{} \\[4pt]
\textbf{Output:} \textless{}User Response\textgreater{} \\
\bottomrule
\end{tabular}
\captionof{table}{The user simulator prompt.}
\label{tab:prompt_simulator}
\end{center}
\vspace{-8pt}

\begin{center}
\begin{tabular}{p{0.92\linewidth}}
\toprule
\textit{\textbf{The slot filling prompt}} \\
\midrule
Your task is to extract and fill information into the schema based on the user response and conversation history. \\[4pt]
\textbf{Inputs:} \\
\quad \textless{}User Response\textgreater{} \\
\quad \textless{}Conversation History\textgreater{} \\[4pt]
\textbf{Output:} \textless{}Updated Schema with Filled Slots\textgreater{} \\
\bottomrule
\end{tabular}
\captionof{table}{The slot filling prompt.}
\label{tab:prompt_slot}
\end{center}
\vspace{-8pt}

\begin{center}
\begin{tabular}{p{0.92\linewidth}}
\toprule
\textit{\textbf{The expertise estimation prompt}} \\
\midrule
Your task is to estimate the user's expertise level based on the collected information. \\[4pt]
\textbf{Inputs:} \\
\quad \textless{}Filled Schema\textgreater{} \\
\quad \textless{}Conversation History\textgreater{} \\[4pt]
\textbf{Output:} \textless{}Estimated Expertise Level\textgreater{} \\
\bottomrule
\end{tabular}
\captionof{table}{The expertise estimation prompt.}
\label{tab:prompt_estimate}
\end{center}
\vspace{-8pt}

\begin{center}
\begin{tabular}{p{0.92\linewidth}}
\toprule
\textit{\textbf{The response generation prompt}} \\
\midrule
Your task is to provide the final answer to the user's initial question. \\[4pt]
\textbf{Inputs:} \\
\quad \textless{}Filled Schema\textgreater{} \\
\quad \textless{}Conversation History\textgreater{} \\
\quad \textless{}Estimated Expertise Level\textgreater{} \\[4pt]
\textbf{Output:} \textless{}Final Answer\textgreater{} \\
\bottomrule
\end{tabular}
\captionof{table}{The response generation prompt.}
\label{tab:prompt_response}
\end{center}
\vspace{-8pt}

\end{document}